\documentclass{article}

\usepackage{times}
\usepackage{graphicx} 
\usepackage{setspace}    
\usepackage[intlimits]{amsmath} 
\usepackage{amsxtra}     
\usepackage{amssymb}
\usepackage{natbib}
\usepackage{algorithm}
\usepackage{algorithmic}
\usepackage{hyperref}

\usepackage{url}

\usepackage[accepted]{icml2016}

\newcommand{\cutsectionup}{\vspace*{-0.08in}}
\newcommand{\cutsectiondown}{\vspace*{-0.05in}}

\newcommand{\cutsubsectionup}{\vspace*{-0.07in}}
\newcommand{\cutsubsectiondown}{\vspace*{-0.05in}}

\icmltitlerunning{Generative Adversarial Text to Image Synthesis}

\begin{document} 

\twocolumn[
\icmltitle{Generative Adversarial Text to Image Synthesis}

\icmlauthor{Scott Reed, Zeynep Akata, Xinchen Yan, Lajanugen Logeswaran}{reedscot$^{1}$, akata$^{2}$, xcyan$^{1}$, llajan$^{1}$}
\icmlauthor{Bernt Schiele, Honglak Lee}{schiele$^{2}$,honglak$^{1}$}
\icmladdress{$^{1}$ University of Michigan, Ann Arbor, MI, USA (\textsc{umich.edu})\\
 $^{2}$ Max Planck Institute for Informatics, Saarbr{\"u}cken, Germany (\textsc{mpi-inf.mpg.de})}
\icmlkeywords{GAN, image generation, vision and language}

\vskip 0.3in
]
\begin{abstract} 
%
Automatic synthesis of realistic images from text would be interesting and useful, but current AI systems are still far from this goal.
However, in recent years generic and powerful recurrent neural network architectures have been developed to learn discriminative text feature representations.
%
Meanwhile, deep convolutional generative adversarial networks (GANs) have begun to generate highly compelling images of specific categories, such as faces, album covers, and room interiors.
In this work, we develop a novel deep architecture and GAN formulation to effectively bridge these advances in text and image modeling, translating visual concepts from characters to pixels.
We demonstrate the capability of our model to generate plausible images of birds and flowers from detailed text descriptions. 
%
%
\vspace*{-0.15in}
\end{abstract} 

\cutsectionup
\section{Introduction}
\label{intro}
\vspace*{-0.05in}
In this work we are interested in translating text in the form of single-sentence human-written descriptions directly into image pixels.
For example, ``this small bird has a short, pointy orange beak and white belly'' or ''the petals of this flower are pink and the anther are yellow''.
The problem of generating images from visual descriptions gained interest in the research community, but it is far from being solved.

Traditionally this type of detailed visual information about an object has been captured in attribute representations - distinguishing characteristics the object category encoded into a vector~\cite{farhadi2009describing,kumar2009attribute,parikh2011relative,LNH13}, in particular to enable zero-shot visual recognition~\cite{FHXFG14,ARWLS15}, and recently for conditional image generation~\cite{yan2015attribute2image}.
%

\begin{figure}[t]
  \begin{center}
    \includegraphics[width=0.44\textwidth]{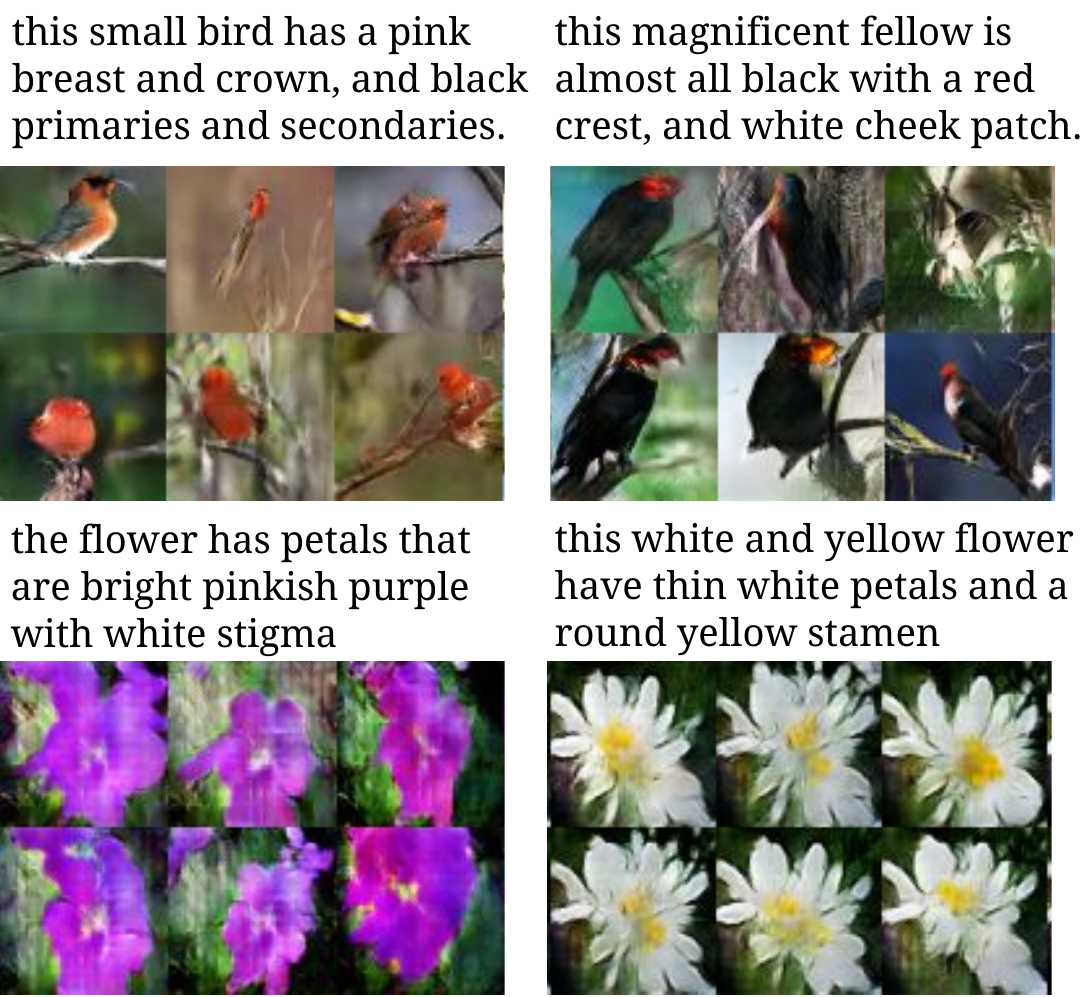}
    \vspace{-0.1in}
    \caption{Examples of generated images from text descriptions. Left: captions are from zero-shot (held out) categories, unseen text. Right: captions are from the training set.}
  \end{center}
  \label{fig:demo_txt2im}
  \vspace{-0.3in}
\end{figure}

While the discriminative power and strong generalization properties of attribute representations are attractive, attributes are also cumbersome to obtain as they may require domain-specific knowledge.
In comparison, natural language offers a general and flexible interface for describing objects in any space of visual categories.
Ideally, we could have the generality of text descriptions with the discriminative power of attributes.

Recently, deep convolutional and recurrent networks for text have yielded highly discriminative and generalizable (in the zero-shot learning sense) text representations learned automatically from words and characters~\cite{reed2015learning}.
These approaches exceed the previous state-of-the-art using attributes for zero-shot visual recognition on the Caltech-UCSD birds database~\cite{wah2011caltech}, and also are capable of zero-shot caption-based retrieval.
Motivated by these works, we aim to learn a mapping directly from words and characters to image pixels.

To solve this challenging problem requires solving two sub-problems: first, learn a text feature representation that captures the important visual details; and second, use these features to synthesize a compelling image that a human might mistake for real.
Fortunately, deep learning has enabled enormous progress in both subproblems - natural language representation and image synthesis - in the previous several years, and we build on this for our current task.

However, one difficult remaining issue not solved by deep learning alone is that the distribution of images conditioned on a text description is highly multimodal, in the sense that there are very many plausible configurations of pixels that correctly illustrate the description.
The reverse direction (image to text) also suffers from this problem but learning is made practical by the fact that the word or character sequence can be decomposed sequentially according to the chain rule; i.e. one trains the model to predict the next token conditioned on the image and all previous tokens, which is a more well-defined prediction problem.

This conditional multi-modality is thus a very natural application for generative adversarial networks~\cite{goodfellow2014generative}, in which the generator network is optimized to fool the adversarially-trained discriminator into predicting that synthetic images are real.
By conditioning both generator and discriminator on side information (also studied by~\citet{mirza2014conditional} and~\citet{denton2015deep}), we can naturally model this phenomenon since the discriminator network acts as a ``smart'' adaptive loss function.

Our main contribution in this work is to develop a simple and effective GAN architecture and training strategy that enables compelling text to image synthesis of bird and flower images from human-written descriptions.
%
We mainly use the Caltech-UCSD Birds dataset and the Oxford-102 Flowers dataset along with five text descriptions per image we collected as our evaluation setting.
Our model is trained on a subset of training categories, and we demonstrate its performance both on the training set categories and on the testing set, i.e. ``zero-shot'' text to image synthesis.
%
%
In addition to birds and flowers, we apply our model to more general images and text descriptions in the MS COCO dataset~\cite{lin2014microsoft}.
%

\cutsectionup
\section{Related work}
\label{related}
\cutsectiondown
%
%
Key challenges in multimodal learning include learning a shared representation across modalities, and to predict missing data (e.g. by retrieval or synthesis) in one modality conditioned on another.
~\citet{ngiam2011multimodal} trained a stacked multimodal autoencoder on audio and video signals and were able to learn a shared modality-invariant representation. 
~\citet{srivastava2012multimodal} developed a deep Boltzmann machine and jointly modeled images and text tags. 
%
%
~\citet{sohn2014improved} proposed a multimodal conditional prediction framework (hallucinating one modality given the other) and provided theoretical justification.

Many researchers have recently exploited the capability of deep convolutional \emph{decoder} networks to generate realistic images.
~\citet{dosovitskiy2015learning} trained a deconvolutional network (several layers of convolution and upsampling) to generate 3D chair renderings conditioned on a set of graphics codes indicating shape, position and lighting.
~\citet{yang2015weakly} added an encoder network as well as actions to this approach. They trained a recurrent convolutional encoder-decoder that rotated 3D chair models and human faces conditioned on action sequences of rotations.
~\citet{reed2015deep} encode transformations from analogy pairs, and use a convolutional decoder to predict visual analogies on shapes, video game characters and 3D cars.

Generative adversarial networks~\cite{goodfellow2014generative} have also benefited from convolutional decoder networks, for the generator network module.
~\citet{denton2015deep} used a Laplacian pyramid of adversarial generator and discriminators to synthesize images at multiple resolutions.
This work generated compelling high-resolution images and could also condition on class labels for controllable generation.
~\citet{radford2015unsupervised} used a standard convolutional decoder, but developed a highly effective and stable architecture incorporating batch normalization to achieve striking image synthesis results.
%

%
The main distinction of our work from the conditional GANs described above is that our model conditions on \emph{text descriptions} instead of class labels.
To our knowledge it is the first end-to-end differentiable architecture from the character level to pixel level.
Furthermore, we introduce a manifold interpolation regularizer for the GAN generator that significantly improves the quality of generated samples, including on held out zero shot categories on CUB. 

The bulk of previous work on multimodal learning from images and text uses retrieval as the target task, i.e. fetch relevant images given a text query or vice versa.
However, in the past year, there has been a breakthrough in using recurrent neural network decoders to generate text descriptions conditioned on images~\citep{vinyals2015show,mao2014deep,KL15,DHGRVSD15}.
These typically condition a Long Short-Term Memory~\cite{hochreiter1997long} on the top-layer features of a deep convolutional network to generate captions using the MS COCO~\cite{lin2014microsoft} and other captioned image datasets.
~\citet{xu2015show} incorporated a recurrent visual attention mechanism for improved results.

Other tasks besides conditional generation have been considered in recent work.
~\citet{ren2015exploring} generate answers to questions about the visual content of images.
This approach was extended to incorporate an explicit knowledge base~\cite{wang2015explicit}.
\citet{zhu2015aligning} applied sequence models to both text (in the form of books) and movies to perform a joint alignment.

In contemporary work~\citet{mansimov2015generating} generated images from text captions, using a variational recurrent autoencoder with attention to paint the image in multiple steps, similar to DRAW~\cite{gregor2015draw}.
Impressively, the model can perform reasonable synthesis of completely novel (unlikely for a human to write) text such as ``a stop sign is flying in blue skies'', suggesting that it does not simply memorize.
%
While the results are encouraging, the problem is highly challenging and the generated images are not yet realistic, i.e., mistakeable for real.
Our model can in many cases generate visually-plausible $64 \times 64$ images conditioned on text, and is also distinct in that our entire model is a GAN, rather only using GAN for post-processing.

Building on ideas from these many previous works, we develop a simple and effective approach for text-based image synthesis using a character-level text encoder and class-conditional GAN.
We propose a novel architecture and learning strategy that leads to compelling visual results.
We focus on the case of fine-grained image datasets, for which we use the recently collected descriptions for Caltech-UCSD Birds and Oxford Flowers with 5 human-generated captions per image~\cite{reed2015learning}.
We train and test on class-disjoint sets, so that test performance can give a strong indication of generalization ability which we also demonstrate on MS COCO images with multiple objects and various backgrounds.
\cutsectionup
\section{Background}
\cutsectiondown
In this section we briefly describe several previous works that our method is built upon. 
%
%
\cutsubsectionup
\subsection{Generative adversarial networks}
\label{preliminary_gan}
\cutsubsectiondown
Generative adversarial networks (GANs) consist of a generator $G$ and a discriminator $D$ that compete in a two-player minimax game: The discriminator tries to distinguish real training data from synthetic images, and the generator tries to fool the discriminator.
Concretely, $D$ and $G$ play the following game on V(D,G):
\begin{align}
\underset{G}{\text{ min }} \underset{D}{\text{ max }} V( D, G) =& \text{ }\mathbb{E}_{ x \sim p_{data}(x)}[\log D(x)] + \\
& \text{ }\mathbb{E}_{ x \sim p_z(z)}[\log(1-D(G(z)))] \nonumber
\end{align}
\citet{goodfellow2014generative} prove that this minimax game has a global optimium precisely when $p_g = p_{data}$, and that under mild conditions (e.g. $G$ and $D$ have enough capacity) $p_g$ converges to $p_{data}$. 
In practice, in the start of training samples from $D$ are extremely poor and rejected by $D$ with high confidence.
It has been found to work better in practice for the generator to maximize $\log(D(G(z)))$ instead of minimizing $\log(1-D(G(z)))$.
%
\cutsubsectionup
\subsection{Deep symmetric structured joint embedding}
\label{preliminary_sje}
\cutsubsectiondown
To obtain a visually-discriminative vector representation of text descriptions, we follow the approach of~\citet{reed2015learning} by using deep convolutional and recurrent text encoders that learn a correspondence function with images.
The text classifier induced by the learned correspondence function $f_t$ is trained by optimizing the following structured loss:
\begin{align}
\label{eq:objective}
\dfrac{1}{N}\sum_{n=1}^{N}\Delta(y_{n},f_v(v_n)) + \Delta(y_{n},f_t(t_n))
\end{align}
where $\{(v_{n},t_{n},y_{n}): n = 1, ..., N\}$ is the training data set, $\Delta$ is the 0-1 loss, $v_n$ are the images, $t_n$ are the corresponding text descriptions, and $y_n$ are the class labels.
Classifiers $f_v$ and $f_t$ are parametrized as follows:
\begin{align}
\label{eq:classifiers}
f_v(v) = \underset{y \in \mathcal{Y}}{\text{arg max }} \mathbb{E}_{t \sim \mathcal{T}(y)}[\phi(v)^T\varphi(t))]\\
f_t(t) = \underset{y \in \mathcal{Y}}{\text{arg max }} \mathbb{E}_{v \sim \mathcal{V}(y)}[\phi(v)^T\varphi(t))]
\end{align}
where $\phi$ is the image encoder (e.g. a deep convolutional neural network), $\varphi$ is the text encoder (e.g. a character-level CNN or LSTM), $\mathcal{T}(y)$ is the set of text descriptions of class $y$ and likewise $\mathcal{V}(y)$ for images.
The intuition here is that a text encoding should have a higher compatibility score with images of the correspondong class compared to any other class and vice-versa. 

To train the model a surrogate objective related to \autoref{eq:objective} is minimized (see~\citet{ARWLS15} for details).
The resulting gradients are backpropagated through $\varphi$ to learn a discriminative text encoder. 
~\citet{reed2015learning} found that different text encoders worked better for CUB versus Flowers, but for full generality and robustness to typos and large vocabulary, in this work we always used a hybrid character-level convolutional-recurrent network.
%

%
%
\cutsectionup
\section{Method}
\label{method}
\cutsectiondown
Our approach is to train a deep convolutional generative adversarial network (DC-GAN) conditioned on text features encoded by a hybrid character-level convolutional-recurrent neural network.
Both the generator network $G$ and the discriminator network $D$ perform feed-forward inference conditioned on the text feature.
\begin{figure*}[ht!]
  \centering
    \includegraphics[width=\textwidth]{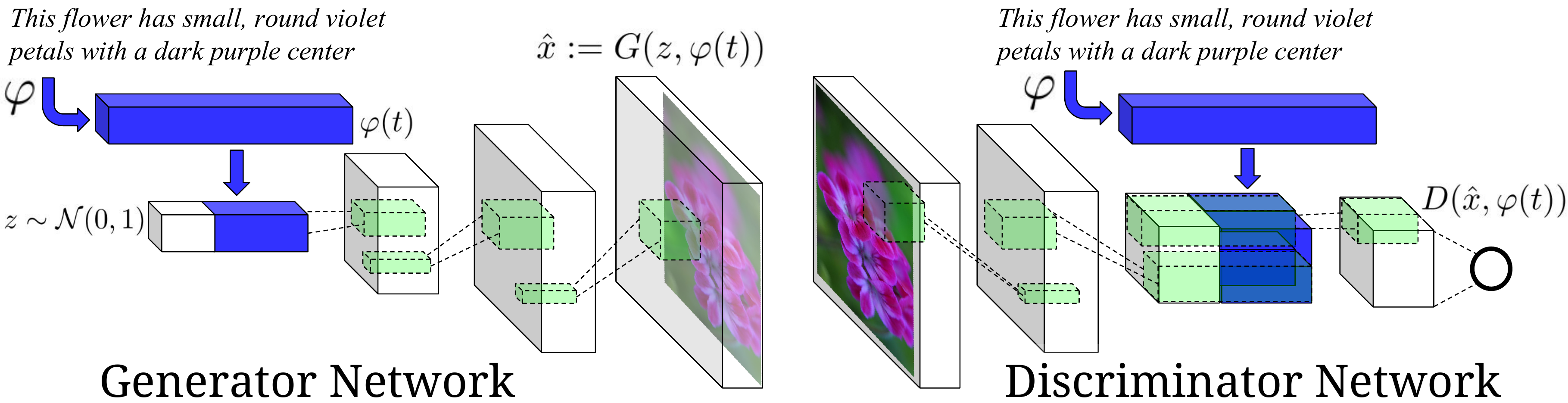}
    \vspace{-0.2in}
    \caption{Our text-conditional convolutional GAN architecture. Text encoding $\varphi(t)$ is used by both generator and discriminator. It is projected to a lower-dimensions and depth concatenated with image feature maps for further stages of convolutional processing.}\label{fig:network} 
  \vspace{-0.1in}
\end{figure*}
\cutsubsectionup
\subsection{Network architecture}
\label{subsec:architecture}
\cutsubsectiondown
We use the following notation.
The generator network is denoted $G : \mathbb{R}^{Z} \times \mathbb{R}^{T} \rightarrow \mathbb{R}^{D}$, the discriminator as $D : \mathbb{R}^{D} \times \mathbb{R}^{T} \rightarrow \{0,1\}$, where $T$ is the dimension of the text description embedding, $D$ is the dimension of the image, and $Z$ is the dimension of the noise input to $G$.
We illustrate our network architecture in \autoref{fig:network}.

In the generator $G$, first we sample from the noise prior $z \in \mathbb{R}^{Z} \sim \mathcal{N}(0,1)$ and we encode the text query $t$ using text encoder $\varphi$.
The description embedding $\varphi(t)$ is first compressed using a fully-connected layer to a small dimension (in practice we used 128) followed by leaky-ReLU and then concatenated to the noise vector $z$.
Following this, inference proceeds as in a normal deconvolutional network: we feed-forward it through the generator $G$; a synthetic image $\hat{x}$ is generated via $\hat{x} \leftarrow G(z, \varphi(t))$.
Image generation corresponds to feed-forward inference in the generator $G$ conditioned on query text and a noise sample.

In the discriminator $D$, we perform several layers of stride-2 convolution with spatial batch normalization~\cite{ioffe2015batch} followed by leaky ReLU.
We again reduce the dimensionality of the description embedding $\varphi(t)$ in a (separate) fully-connected layer followed by rectification.
When the spatial dimension of the discriminator is $4 \times 4$, we replicate the description embedding spatially and perform a depth concatenation.
We then perform a $1 \times 1$ convolution followed by rectification and a $4 \times 4$ convolution to compute the final score from $D$.
Batch normalization is performed on all convolutional layers.

\cutsubsectionup
\subsection{Matching-aware discriminator (GAN-CLS)}
\label{subsec:GAN-CLS}
\cutsubsectiondown

The most straightforward way to train a conditional GAN is to view (text, image) pairs as joint observations and train the discriminator to judge pairs as real or fake.
This type of conditioning is naive in the sense that the discriminator has no explicit notion of whether real training images match the text embedding context.

However, as discussed also by~\cite{gauthierconditional}, the dynamics of learning may be different from the non-conditional case.
In the beginning of training, the discriminator ignores the conditioning information and easily rejects samples from $G$ because they do not look plausible.
Once $G$ has learned to generate plausible images, it must also learn to align them with the conditioning information, and likewise $D$ must learn to evaluate whether samples from $G$ meet this conditioning constraint.

In naive GAN, the discriminator observes two kinds of inputs: real images with matching text, and synthetic images with arbitrary text.
Therefore, it must implicitly separate two sources of error: unrealistic images (for \emph{any} text), and realistic images of the wrong class that mismatch the conditioning information. 
Based on the intuition that this may complicate learning dynamics, we modified the GAN training algorithm to separate these error sources.
In addition to the real / fake inputs to the discriminator during training, we add a third type of input consisting of real images with mismatched text, which the discriminator must learn to score as fake.
By learning to optimize image / text matching in addition to the image realism, the discriminator can provide an additional signal to the generator.
\setlength{\textfloatsep}{8pt}
\begin{algorithm}[t]
	\begin{algorithmic}[1]
		\STATE {\bfseries Input:} minibatch images $x$, matching text $t$, mismatching $\hat{t}$, number of training batch steps $S$
		\FOR{$n=1$ {\bfseries to} $S$}
		\STATE $h \gets \varphi(t)$ \COMMENT{Encode matching text description}
		\STATE $\hat{h} \gets \varphi(\hat{t})$ \COMMENT{Encode mis-matching text description}
		\STATE $z \sim \mathcal{N}(0,1)^{Z}$ \COMMENT{Draw sample of random noise}
		\STATE $\hat{x} \gets G(z, h)$ \COMMENT{Forward through generator}
		
		\STATE $s_{r} \gets D(x, h)$ \COMMENT{real image, right text}
		\STATE $s_{w} \gets D(x, \hat{h})$ \COMMENT{real image, wrong text}
		\STATE $s_{f} \gets D(\hat{x}, h)$ \COMMENT{fake image, right text}
		\STATE $\mathcal{L}_{D} \gets \log(s_{r}) + (\log(1-s_{w}) + \log(1-s_{f})) / 2$
		\STATE $D \gets D - \alpha \partial{\mathcal{L}}_{D} / \partial D$ \COMMENT{Update discriminator}
		\STATE $\mathcal{L}_G \gets \log(s_{f})$
		\STATE $G \gets G - \alpha \partial{\mathcal{L}}_G / \partial G$ \COMMENT{Update generator}
		\ENDFOR
	\end{algorithmic}
	\caption{GAN-CLS training algorithm with step size $\alpha$, using minibatch SGD for simplicity.\label{alg:training}}
\end{algorithm}

Algorithm~\autoref{alg:training} summarizes the training procedure. 
After encoding the text, image and noise (lines 3-5) we generate the fake image ($\hat{x}$, line 6).
$s_{r}$ indicates the score of associating a real image and its corresponding sentence (line 7), $s_{w}$ measures the score of associating a real image with an arbitrary sentence (line 8), and $s_{f}$ is the score of associating a fake image with its corresponding text (line 9).
Note that we use $\partial \mathcal{L}_{D} / \partial D$ to indicate the gradient of $D$'s objective with respect to its parameters, and likewise for $G$.
Lines 11 and 13 are meant to indicate taking a gradient step to update network parameters.
\cutsubsectionup
\subsection{Learning with manifold interpolation (GAN-INT)}
\label{subsec:GAN-INT}
\cutsubsectiondown
%
%
Deep networks have been shown to learn representations in which interpolations between embedding pairs tend to be near the data manifold~\cite{bengio2013better,reed2014learning}.
Motivated by this property, we can generate a large amount of additional text embeddings by simply interpolating between embeddings of training set captions.
Critically, these interpolated text embeddings need not correspond to any actual human-written text, so there is no additional labeling cost.
This can be viewed as adding an additional term to the generator objective to minimize:
\begin{align}
\mathbb{E}_{t_1,t_2 \sim p_{data}}[ \log (1 - D(G(z, \beta t_1 + (1-\beta) t_2 ))) ]
\end{align}
where $z$ is drawn from the noise distribution and $\beta$ interpolates between text embeddings $t_1$ and $t_2$.
In practice we found that fixing $\beta = 0.5$ works well.

Because the interpolated embeddings are synthetic, the discriminator $D$ does not have ``real'' corresponding image and text pairs to train on.
However, $D$ learns to predict whether image and text pairs match or not.
Thus, if $D$ does a good job at this, then by satisfying $D$ on interpolated text embeddings $G$ can learn to fill in gaps on the data manifold in between training points.
Note that $t_1$ and $t_2$ may come from different images and even different categories.\footnote{In our experiments, we used fine-grained categories (e.g. birds are similar enough to other birds, flowers to other flowers, etc.), and interpolating across categories did not pose a problem.}

\cutsubsectionup
\subsection{Inverting the generator for style transfer}
\label{subsec:style}
\cutsubsectiondown
If the text encoding $\varphi(t)$ captures the image content (e.g. flower shape and colors), then in order to generate a realistic image the noise sample $z$ should capture style factors such as background color and pose.
With a trained GAN, one may wish to transfer the style of a query image onto the content of a particular text description.
To achieve this, one can train a convolutional network to invert $G$ to regress from samples $\hat{x} \leftarrow G(z,\varphi(t))$ back onto $z$.
%
%
We used a simple squared loss to train the style encoder:
\begin{align}
\mathcal{L}_{style} = \mathbb{E}_{t, z \sim \mathcal{N}(0,1)} || z - S(G(z,\varphi(t))) ||^{2}_{2}
\end{align}
where $S$ is the style encoder network.
With a trained generator and style encoder, style transfer from a query image $x$ onto text $t$ proceeds as follows:
\begin{align}
s \leftarrow S(x)\text{,   }
\hat{x} \leftarrow G(s, \varphi(t))\nonumber
\end{align}
where $\hat{x}$ is the result image and $s$ is the predicted style.
\cutsectionup
\section{Experiments} 
\label{experiments}
\cutsectiondown
\begin{figure*}[t!]
  \begin{center}
    \includegraphics[width=\textwidth]{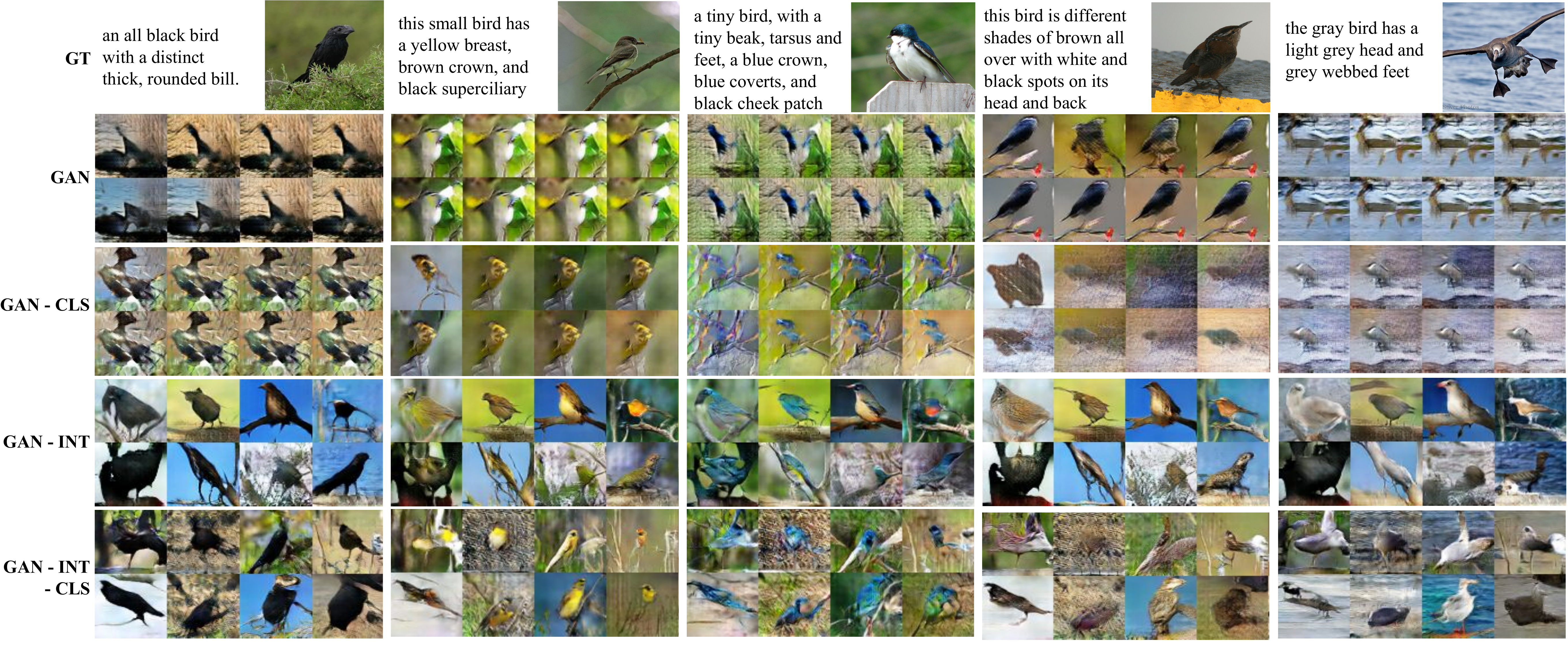}
  \end{center}
  \vspace{-0.2in}
  \caption{Zero-shot (i.e. conditioned on text from unseen test set categories) generated bird images using GAN, GAN-CLS, GAN-INT and GAN-INT-CLS. We found that interpolation regularizer was needed to reliably achieve visually-plausible results.}
  \label{fig:cub_qualitative}
\end{figure*}
\begin{figure*}[t!]
  \begin{center}
    \includegraphics[width=\textwidth]{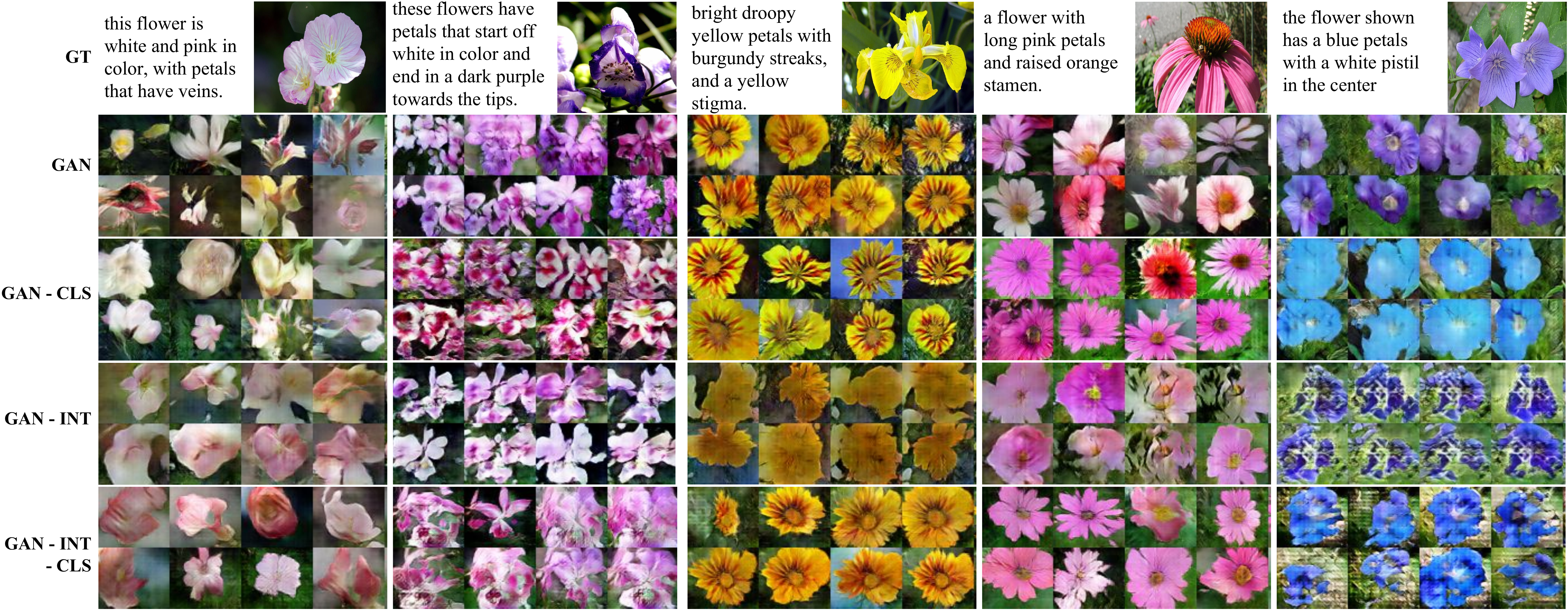}
  \end{center}
  \vspace{-0.2in}
  \caption{Zero-shot generated flower images using GAN, GAN-CLS, GAN-INT and GAN-INT-CLS. All variants generated plausible images. Although some shapes of test categories were not seen during training (e.g. columns 3 and 4), the color information is preserved.}
  \label{fig:flowers_qualitative}
    \vspace{-0.1in}
\end{figure*}
In this section we first present results on the CUB dataset of bird images and the Oxford-102 dataset of flower images.
CUB has 11,788 images of birds belonging to one of 200 different categories.
The Oxford-102 contains 8,189 images of flowers from 102 different categories.

As in~\citet{ARWLS15} and~\citet{reed2015learning}, we split these into class-disjoint training and test sets.
CUB has 150 train+val classes and 50 test classes, while Oxford-102 has 82 train+val and 20 test classes.
For both datasets, we used 5 captions per image.
During mini-batch selection for training we randomly pick an image view (e.g. crop, flip) of the image and one of the captions. 

For text features, we first pre-train a deep convolutional-recurrent text encoder on structured joint embedding of text captions with 1,024-dimensional GoogLeNet image embedings~\cite{szegedy2015going} as described in \autoref{preliminary_sje}.
For both Oxford-102 and CUB we used a hybrid of character-level ConvNet with a recurrent neural network (char-CNN-RNN) as described in~\cite{reed2015learning}.
Note, however that pre-training the text encoder is not a requirement of our method and we include some end-to-end results in the supplement.
The reason for pre-training the text encoder was to increase the speed of training the other components for faster experimentation.
We also provide some qualitative results obtained with MS COCO images of the validation set to show the generalizability of our approach.

We used the same GAN architecture for all datasets.
The training image size was set to $64 \times 64 \times 3$.
The text encoder produced $1,024$-dimensional embeddings that were projected to $128$ dimensions in both the generator and discriminator before depth concatenation into convolutional feature maps.

As indicated in Algorithm~\autoref{alg:training}, we take alternating steps of updating the generator and the discriminator network.
We used the same base learning rate of $0.0002$, and used the ADAM solver~\cite{Ba_ICLR_2015} with momentum $0.5$.
The generator noise was sampled from a $100$-dimensional unit normal distribution.
We used a minibatch size of $64$ and trained for $600$ epochs.
Our implementation was built on top of \texttt{dcgan.torch}\footnote{\url{https://github.com/soumith/dcgan.torch}}.
\cutsubsectionup
\subsection{Qualitative results}
\cutsubsectiondown
We compare the GAN baseline, our GAN-CLS with image-text matching discriminator (\autoref{subsec:GAN-CLS}), GAN-INT learned with text manifold interpolation (\autoref{subsec:GAN-INT}) and GAN-INT-CLS which combines both.

Results on CUB can be seen in \autoref{fig:cub_qualitative}.
%
%
GAN and GAN-CLS get some color information right, but the images do not look real.
However, GAN-INT and GAN-INT-CLS show plausible images that usually match all or at least part of the caption. 
We include additional analysis on the robustness of each GAN variant on the CUB dataset in the supplement.

Results on the Oxford-102 Flowers dataset can be seen in \autoref{fig:flowers_qualitative}.
In this case, all four methods can generate plausible flower images that match the description.
The basic GAN tends to have the most variety in flower morphology (i.e. one can see very different petal types if this part is left unspecified by the caption), while other methods tend to generate more class-consistent images.
We speculate that it is easier to generate flowers, perhaps because birds have stronger structural regularities across species that make it easier for $D$ to spot a fake bird than to spot a fake flower.

Many additional results with GAN-INT and GAN-INT-CLS as well as GAN-E2E (our end-to-end GAN-INT-CLS without pre-training the text encoder $\varphi(t)$) for both CUB and Oxford-102 can be found in the supplement.
\cutsubsectionup
\subsection{Disentangling style and content}
\cutsubsectiondown
In this section we investigate the extent to which our model can separate style and content.
By content, we mean the visual attributes of the bird itself, such as shape, size and color of each body part.
By style, we mean all of the other factors of variation in the image such as background color and the pose orientation of the bird.

The text embedding mainly covers content information and typically nothing about style, e.g. captions do not mention the background or the bird pose.
Therefore, in order to generate realistic images then GAN must learn to use noise sample $z$ to account for style variations. 

To quantify the degree of disentangling on CUB we set up two prediction tasks with noise $z$ as the input: pose verification and background color verification.
For each task, we first constructed similar and dissimilar pairs of images and then computed the predicted style vectors by feeding the image into a style encoder (trained to invert the input and output of generator).
If GAN has disentangled style using $z$ from image content, the similarity between images of the same style (e.g. similar pose) should be higher than that of different styles (e.g. different pose).

\begin{figure}
\centering
  \includegraphics[width=\columnwidth]{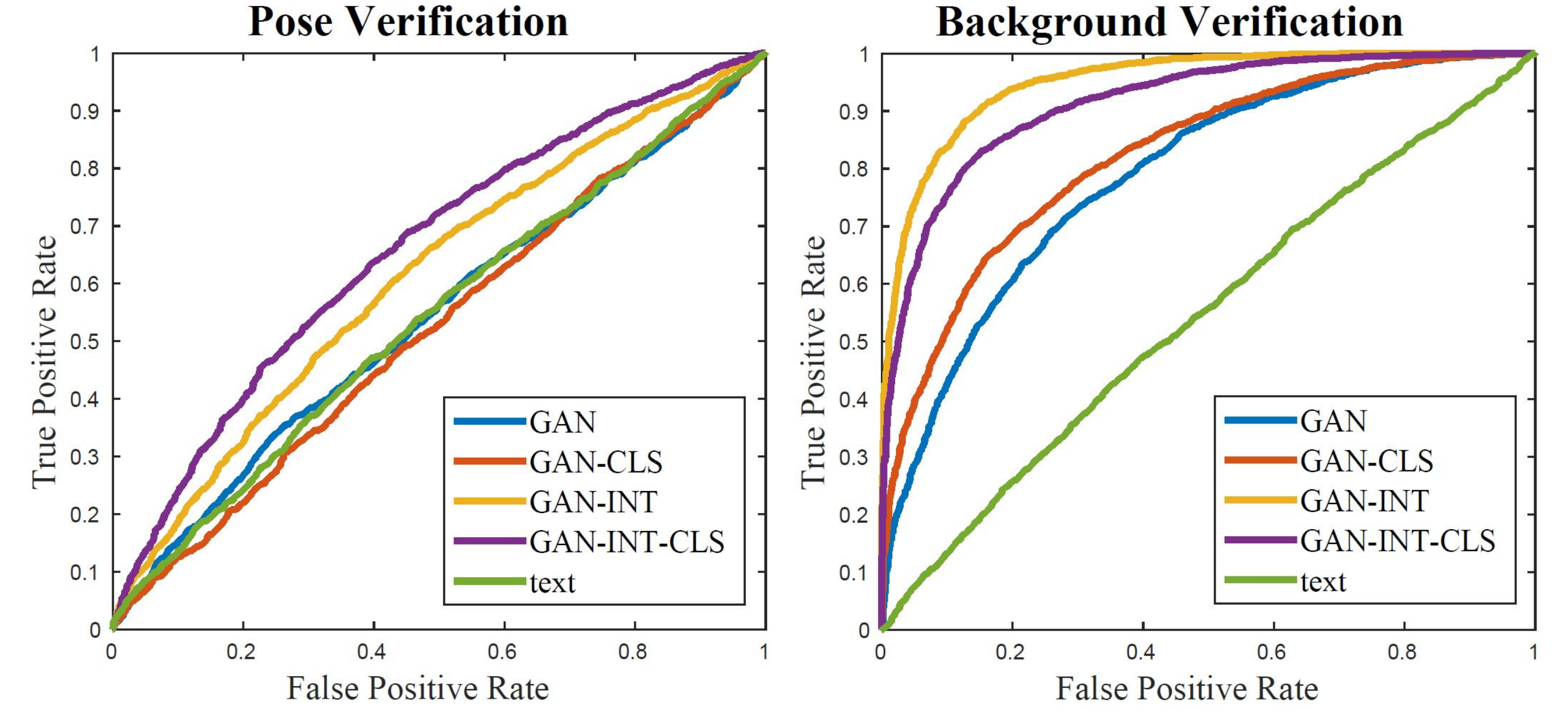}
  \vspace{-0.3in}
  \caption{ROC curves using cosine distance between predicted style vector on same vs. different style image pairs. Left: image pairs reflect same or different pose. Right: image pairs reflect same or different average background color.}
  \label{fig:roc}
   \vspace{-0.0in}
\end{figure}

To recover $z$, we inverted the each generator network as described in \autoref{subsec:style}.
To construct pairs for verification, we grouped images into 100 clusters using K-means where images from the same cluster share the same style.
For background color, we clustered images by the average color (RGB channels) of the background; for bird pose, we clustered images by 6 keypoint coordinates (beak, belly, breast, crown, forehead, and tail).

For evaluation, we compute the actual predicted style variables by feeding pairs of images style encoders for GAN, GAN-CLS, GAN-INT and GAN-INT-CLS.
We verify the score using cosine similarity and report the AU-ROC (averaging over 5 folds).
As a baseline, we also compute cosine similarity between text features from our text encoder.

We present results on \autoref{fig:roc}. As expected, captions alone are not informative for style prediction.
Moreover, consistent with the qualitative results, we found that models incorporating interpolation regularizer (GAN-INT, GAN-INT-CLS) perform the best for this task. 

\cutsubsectionup
\subsection{Pose and background style transfer}
\begin{figure}[t]
	\begin{center}
		\includegraphics[width=\columnwidth]{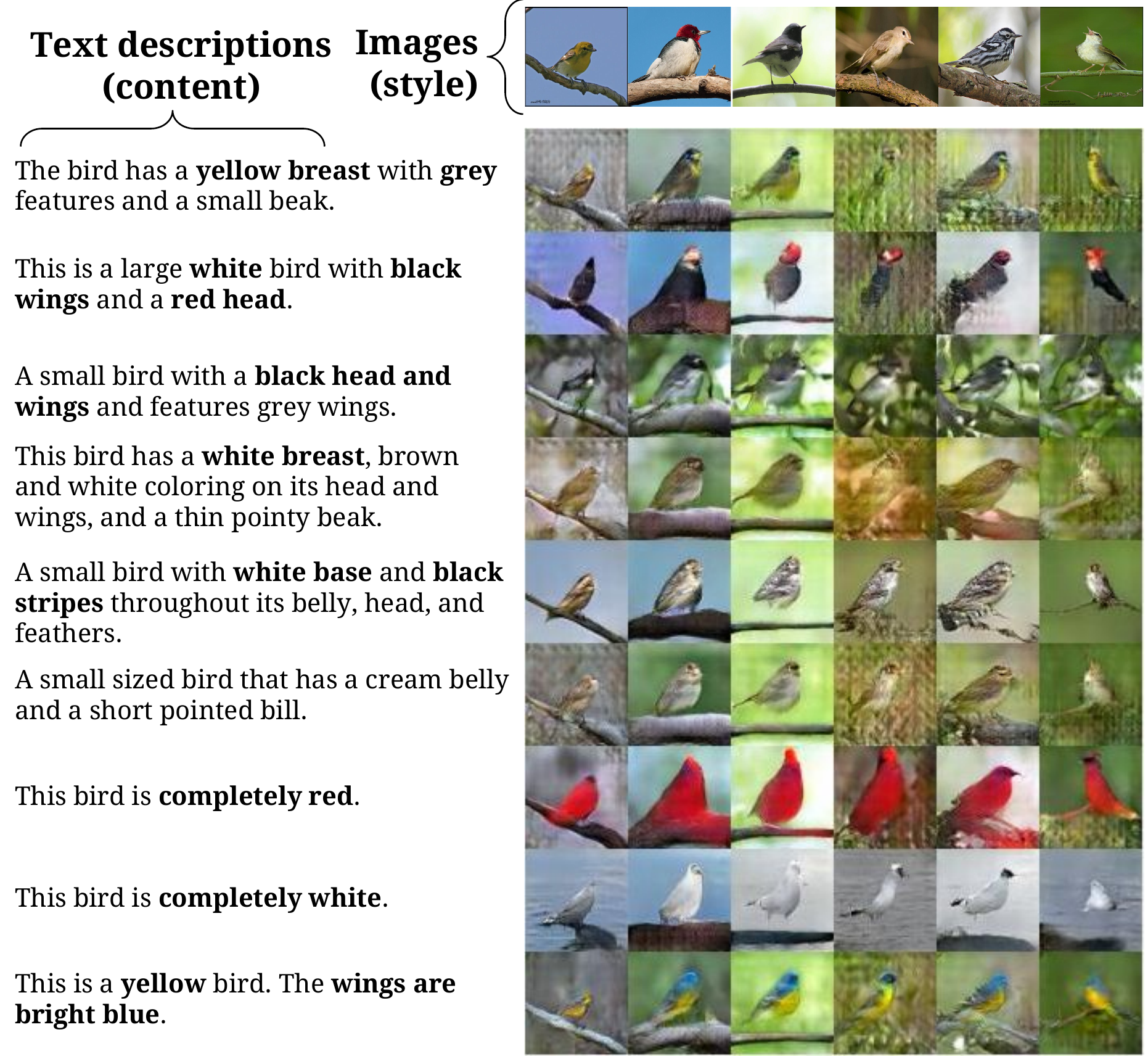}
	\end{center}
	\vspace*{-0.15in}
	\caption{Transfering style from the top row (real) images to the content from the query text, with $G$ acting as a deterministic decoder. The bottom three rows are captions made up by us.}
	\label{fig:cub_style_transfer}
	\vspace*{-0.05in}
\end{figure}

\cutsubsectiondown
We demonstrate that GAN-INT-CLS with trained style encoder (\autoref{subsec:style}) can perform style transfer from an unseen query image onto a text description.
\autoref{fig:cub_style_transfer} shows that images generated using the inferred styles can accurately capture the pose information.
In several cases the style transfer preserves detailed background information such as a tree branch upon which the bird is perched.

Disentangling the style by GAN-INT-CLS is interesting because it suggests a simple way of generalization. This way we can combine previously seen content (e.g. text) and previously seen styles, but in novel pairings so as to generate plausible images very different from any seen image during training.
Another way to generalize is to use attributes that were previously seen (e.g. blue wings, yellow belly) as in the generated parakeet-like bird in the bottom row of \autoref{fig:cub_style_transfer}.
This way of generalization takes advantage of text representations capturing multiple visual aspects.

\begin{figure*}[t]
  \centering
    \includegraphics[width=\textwidth]{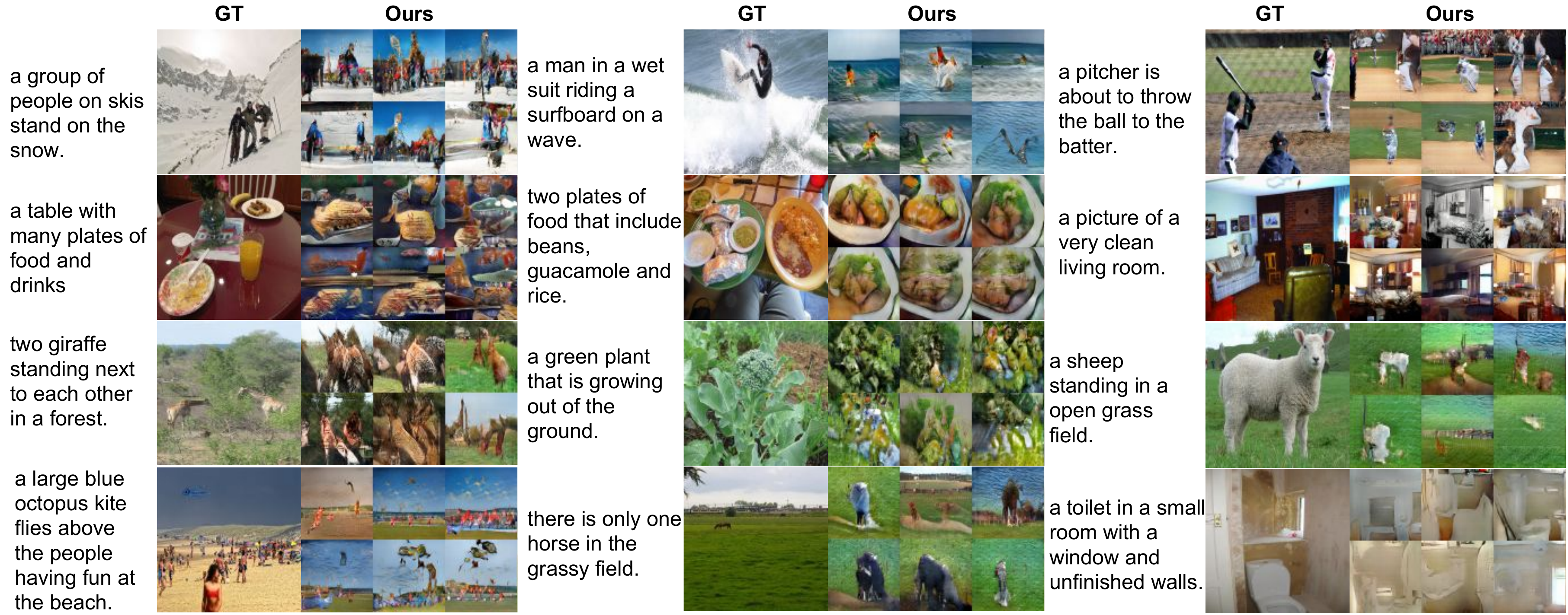}
    \vspace{-0.25in}
    \caption{Generating images of general concepts using our GAN-CLS on the MS-COCO validation set. Unlike the case of CUB and Oxford-102, the network must (try to) handle multiple objects and diverse backgrounds.} 
    \vspace{-0.2in}
    \label{fig:coco_ours}
\end{figure*}
\cutsubsectionup
\subsection{Sentence interpolation} 
\begin{figure}[t]
	\centering
	\includegraphics[width=\columnwidth]{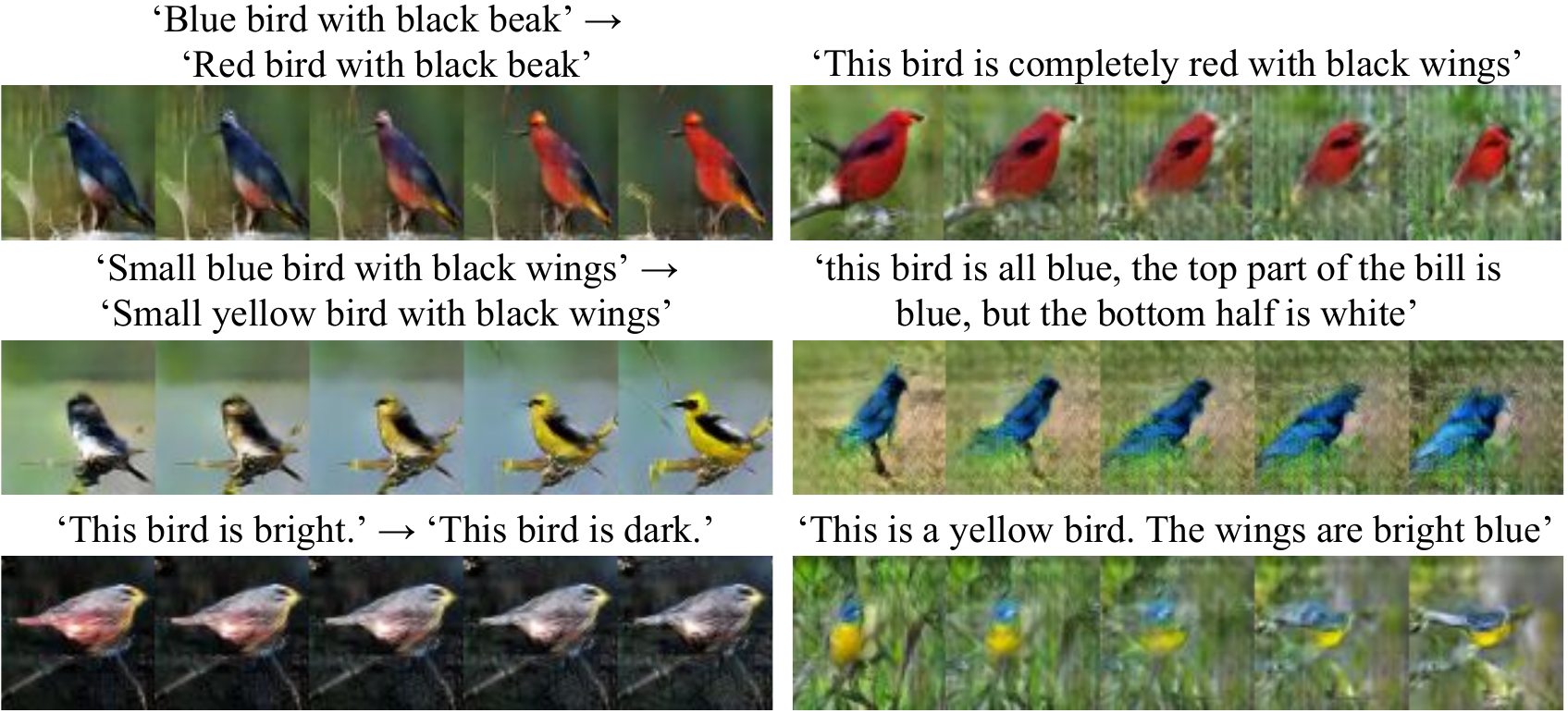}
	\vspace{-0.25in}
	\caption{Left: Generated bird images by interpolating between two sentences (within a row the noise is fixed). Right: Interpolating between two randomly-sampled noise vectors.\label{fig:cub_txt_interp}}
	\vspace*{-0.0in}
\end{figure}

\cutsubsectiondown
%
\autoref{fig:cub_txt_interp} demonstrates the learned text manifold by interpolation (Left).
Although there is no ground-truth text for the intervening points, the generated images appear plausible.
Since we keep the noise distribution the same, the only changing factor within each row is the text embedding that we use. 
Note that interpolations can accurately reflect color information, such as a bird changing from blue to red while the pose and background are invariant.

As well as interpolating between two text encodings, we show results on \autoref{fig:cub_txt_interp} (Right) with noise interpolation.
Here, we sample two random noise vectors. 
By keeping the text encoding fixed, we interpolate between these two noise vectors and generate bird images with a smooth transition between two styles by keeping the content fixed.

\cutsubsectionup
\subsection{Beyond birds and flowers}
\cutsubsectiondown
We trained a GAN-CLS on MS-COCO to show the generalization capability of our approach on a general set of images that contain multiple objects and variable backgrounds.
We use the same text encoder architecture, same GAN architecture and same hyperparameters (learning rate, minibatch size and number of epochs) as in CUB and Oxford-102.  
The only difference in training the text encoder is that COCO does not have a single object category per class.
However, we can still learn an instance level (rather than category level) image and text matching function, as in~\cite{kiros2014unifying}.

Samples and ground truth captions and their corresponding images are shown on \autoref{fig:coco_ours}.
A common property of all the results is the sharpness of the samples, similar to other GAN-based image synthesis models. 
We also observe diversity in the samples by simply drawing multiple noise vectors and using the same fixed text encoding.

From a distance the results are encouraging, but upon close inspection it is clear that the generated scenes are not usually coherent; for example the human-like blobs in the baseball scenes lack clearly articulated parts.
In future work, it may be interesting to incorporate hierarchical structure into the image synthesis model in order to better handle complex multi-object scenes.

%

A qualitative comparison with AlignDRAW~\citep{mansimov2015generating} can be found in the supplement.
%
%
GAN-CLS generates sharper and higher-resolution samples that roughly correspond to the query, but AlignDRAW samples more noticably reflect single-word changes in the selected queries from that work.
%
%
Incorporating temporal structure into the GAN-CLS generator network could potentially improve its ability to capture these text variations.
\cutsectionup
\section{Conclusions}
\label{conclusions}
\cutsectiondown
In this work we developed a simple and effective model for generating images based on detailed visual descriptions.
We demonstrated that the model can synthesize many plausible visual interpretations of a given text caption.
%
%
Our manifold interpolation regularizer substantially improved the text to image synthesis on CUB.
We showed disentangling of style and content, and bird pose and background transfer from query images onto text descriptions.
Finally we demonstrated the generalizability of our approach to generating images with multiple objects and variable backgrounds with our results on MS-COCO dataset.
In future work, we aim to further scale up the model to higher resolution images and add more types of text. 

{

\cutsectionup
\section*{Acknowledgments}
\cutsectiondown
This work was supported in part by NSF CAREER IIS-1453651, ONR N00014-13-1-0762 and NSF CMMI-1266184.
\bibliography{references}
\bibliographystyle{icml2016}
}

\end{document}